\documentclass{BMVCarXiv}

\usepackage{comment}
\usepackage{graphicx}
\usepackage{amsmath,amssymb}
\usepackage{color}
\usepackage{microtype}
\usepackage{soul}
\usepackage{tabularx}
\usepackage{booktabs}
\usepackage{amsmath}
\usepackage{wrapfig}
\usepackage{xcolor}

\newcolumntype{L}{>{\raggedright\let\newline\\\arraybackslash\hspace{0pt}}m{0.7cm}}

\begin{document}

\def\mysection#1#2{\section{#1}\label{sec:#2}}
\def\mysubsection#1#2{\subsection{#1}\label{sec:#2}}
\def\mysubsubsection#1#2{\subsubsection{#1}\label{sec:#2}}
\def\figurePath{images/}
\def\myfigure#1#2{\begin{figure}[htb]\centering\includegraphics*[width = \linewidth]{\figurePath#1}\caption{#2}\label{fig:#1}\end{figure}}

\newcommand{\eg}{e.\,g., }
\newcommand{\ie}{i.\,e., }
\newcommand{\etal}{et~al.\ }
\newcommand{\RGBD}{RGB\=/D\ }
\newcommand{\citeetal}[1]{et~al.~\shortcite{#1}}
\newcommand{\argmax}[1]{\underset{#1}{\operatorname{arg\,max}}}
\newcommand{\argmin}[1]{\underset{#1}{\operatorname{arg\,min}\ }}
\newcommand{\diag}[1]{\operatorname{diag}(#1)}
\newcommand{\refSec}[1]{Sec.~\ref{sec:#1}}
\newcommand{\refFig}[1]{Fig.~\ref{fig:#1}}
\newcommand{\refEq}[1]{Eq.~\ref{eq:#1}}
\newcommand{\refTbl}[1]{Tbl.~\ref{tbl:#1}}
\newcommand{\tblhead}[1]{\multicolumn{1}{c}{#1}}

\newcommand{\unsure}[1]{{\sethlcolor{yellow}\hl{#1}}}

\newcommand{\change}[1]{#1}

\soulregister\ref7
\soulregister\refFig7
\soulregister\cite7
\soulregister\ref7
\soulregister\pageref7
\soulregister\shortcite7
\soulregister\eg0
\soulregister\ie0
\soulregister\etal0

\DeclareGraphicsExtensions{.png,.jpg,.pdf,.ai,.psd}
\DeclareGraphicsRule{.ai}{pdf}{.ai}{}
\DeclareGraphicsRule{.psd}{pdf}{.psd}{}

\title{Learning on the Edge:\\Explicit Boundary Handling in CNNs}


\addauthor{Carlo Innamorati}{c.innamorati@cs.ucl.ac.uk}{1}
\addauthor{Tobias Ritschel}{t.ritschel@cs.ucl.ac.uk}{1}
\addauthor{Tim Weyrich}{t.weyrich@cs.ucl.ac.uk}{1}
\addauthor{Niloy J. Mitra}{n.mitra@cs.ucl.ac.uk}{1}
\addinstitution{University College London}

\runninghead{Innamorati et al.}{Learning on the Edge}

\maketitle

\begin{abstract}%
Convolutional neural networks (CNNs) handle the case where filters extend beyond the image boundary using several heuristics, such as \texttt{zero}, \texttt{repeat} or \texttt{mean} padding. 
These schemes are applied in an ad-hoc fashion and, being weakly related to the image content and oblivious of the target task, result in low output quality at the boundary.
In this paper, we propose a simple and effective improvement that \textit{learns} the boundary handling itself.
At training-time, the network is provided with a separate set of \texttt{explicit} boundary filters.
At testing-time, we use these filters which have learned to extrapolate features at the boundary in an optimal way for the specific task. 
Our extensive evaluation, over a wide range of architectural changes (variations of layers, feature channels, or both), shows how the \texttt{explicit} filters result in improved boundary handling. 
Consequently, we demonstrate an improvement of 5\,\% to 20\,\% across the board of typical CNN applications (colorization, de-Bayering, optical flow, and disparity estimation).

\end{abstract}

\section{Introduction}
When performing convolutions on a finite domain, boundary rules are required as the kernel's support extends beyond the edge.
For convolutional neural networks (CNNs), many discrete filter kernels ``slide'' over a 2D image and typically boundary rules including \texttt{zero}, \texttt{reflect}, \texttt{mean}, \texttt{clamp} are used to extrapolate values outside the image.

\myfigure{Problem}{
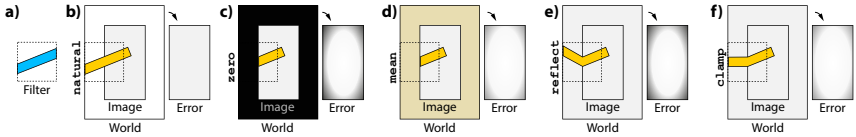
Applying a feature detection-like filter \textbf{(a)} to an image  with different boundary rules \textbf{(b--f)}. 
We show the error as the ratio of the ideal and the observed response.
A bright value means a low error due to a ratio of 1 \ie the response is similar to the ideal condition.
Darker values indicate a deterioration.
} 

Considering a simple detection filter (\refFig{Problem}a) applied to a diagonal feature (\refFig{Problem}b), we see that no boundary rule is ever ideal: \texttt{zero} will create a black boundary halo (\refFig{Problem}c), using the \texttt{mean} color will reduce but not remove the issue (\refFig{Problem}d), \texttt{reflect} and \texttt{clamp} (\refFig{Problem}e and \ref{fig:Problem}f) will create different kinks in a diagonal edge where the ground-truth continuation would be straight.
In \refFig{Problem} we visualize this as the error between the ideal response and the response we would observe at a location if a feature was present.
In practical feature channels, these will manifest as false positive and negative images.
These deteriorate overall feature quality, not only on the boundary but also inside. 
Another, equally unsatisfying, solution is to execute the CNN only on a ``valid'' interior part of the input image (crop), or to execute it multiple times and merge the outcome slide.
Working in lower or multiple resolutions, the problem is even stronger, as low-resolution images have a higher percentage of boundary pixels.
In a typical modern encoder-decoder \cite{ronneberger2015u}, \emph{all} will eventually become boundary pixels at some step.

Having a second thought on what a 2D image actually is, we see, that the ideal boundary rule would be the one that extends the content exactly to the values an image taken with a larger sensor would have contained.
Such a rule appears elusively hard to come by as it relies on information not observed.
We cannot decide with certainty from observing the yellow part inside the image in \refFig{Problem}b how the part outside the image continues -- what if the yellow structure really stopped? -- and therefore it is unknown what the filter response should be.
However, neural networks have the ability to extrapolate information from a context, for example in in-painting tasks \cite{ren2015shepard}.
Here, this context is the image part inside the boundary.
Given this observation, not every extension is equally likely.
Most human observers would follow the Gestalt assumption of continuity and predict the yellow bar to continue at constant slope outside the image.
Can a CNN do this extrapolation while extracting features?

Addressing the boundary challenge, and making use of a CNN's extrapolating power, we propose the use of a novel \texttt{explicit} boundary rule in CNNs.
As such rules will have to depend on the image content and the spatial location of that content, we advocate to model them as a set of learned \emph{boundary filters} that simply replace the non-boundary filters when executed on the boundary.
These boundary filters are supposed to produce exactly the same feature channels the non-boundary filters produce.
Every boundary configuration (upper edge, lower left corner, etc.) has a different filter.
This implies, that they incur no time or space overhead at runtime.
At training-time, boundary and non-boundary filters are jointly optimized and no additional steps are required.

It seems, that introducing more degrees of freedom increases the optimization challenge.
However, introducing the right degrees of freedom, can actually turn an unsolvable problem into separate tasks that have simple independent solutions, as we conclude from a reduction of error both at the interior and at the edges, when using our method.

After reviewing previous work and introducing our formalism,
we demonstrate how using \texttt{explicit} boundary conditions can improve the quality across a wide range of possible architectures (\refSec{Analysis}).
We next show improvement in performance for tasks such as de-noising and de-bayering \cite{gharbi2016deep},  colorization \cite{zhang2016colorful} as well as disparity and scene flow \cite{dosovitskiy2015flownet}, in \refSec{Applications}.  

\section{Previous Work}
Our work extends deep convolutional neural networks \cite{goodfellow2016deep} (CNNs).
To our knowledge, the immediate effect of boundary handling has not been looked into explicitly.
CNNs owe a part of their effectiveness to weight-sharing or shift-invariance property: only a single convolution needs to be optimized that is applied to the entire image \cite{fukushima1982neocognitron}.
Doing so, inevitably, the filter kernel will touch upon the image boundary at some point.
Classic CNNs use zero padding \cite{ciregan2012multi}, \ie they enlarge the image by the filter kernel size they use, or directly  crop, \ie run only on a subset \cite{krizhevsky2012imagenet} and discard the boundary.
Another simple solution is to perform filtering with an arbitrary boundary handling and crop the part of the image that remains unaffected: if the filter is centered and 3 pixels wide, a 100$\times$100 pixel image is cropped to 98$\times$98 pixels.
This works in a single resolution, but multiple layers, in particular at multiple resolutions, grow the region affected by the boundary linearly or even exponentially.
For example, the seminal U-net \cite{ronneberger2015u} employs a complicated sliding scheme to produce central patches from a context that is affected by the boundary, effectively computing a large fraction of values that are never used.
We show how exactly such a U-net-like architecture can be combined with \texttt{explicit} boundaries to realize a better efficacy with lower implementation and runtime overhead. 
Other work has extended the notion of invariance to flips \cite{cohen2016group} and rotations \cite{worrall2017harmonic}.
Our extension could be seen as adding invariance under boundary conditions.
For some tasks like in-panting, however, invariance is not desired, and translation-variant convolutions are used \cite{ren2015shepard}.
This paper shares the idea to use different convolutions in different spatial locations.
Uhrig~\etal have weighted convolutions to skip pixels undefined at test time \cite{uhrig2017sparsity}.
In our setting, the undefined pixels are known at train time to always fall on the boundary.
By making this explicit to the learning, it can capitalize on knowing how the image extends.

\mysection{Explicit Boundary Rules}{OurApproach}
In this section we will define convolutions that can account for \texttt{explicit} boundary rules, before discussing the loss and implementation options.

\paragraph{Convolution}
Key to \texttt{explicit} boundary handling is a domain decomposition.
Intuitively, in our approach, instead of running the same filter for every pixel, different filters are run at the boundary.
In any case, they compute the same feature.
This is done independently for every convolution kernel in the network.
For simplicity, we will here explain the idea for a single kernel that computes a single feature.
The extension to many kernels and features is straightforward.
Again, for simplicity, we describe the procedure for a 2D convolution, mapping scalar input to scalar output.
The 3D convolution, mapping higher-dimensional input to scalar output is derived similarly. 

\myfigure{Domain}{
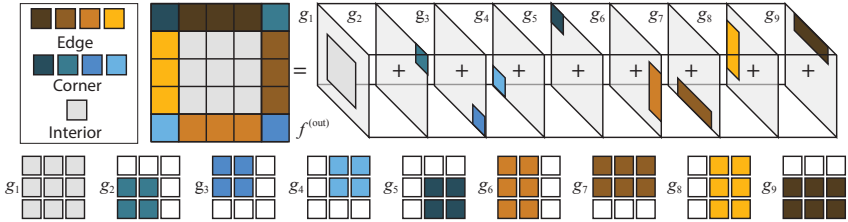
Example domain decomposition for a 5$\times$5 image.
Colors encode different filters.
}

A common \texttt{zero} boundary handling convolution $\ast_0$ of 
an input image $f^{(\mathrm{in})}$ with the kernel $g$ is defined as \begin{align}
f^{(\mathrm{out})}[\mathbf x]\ast_0 g
=
\sum_{\mathbf y \in \mathcal K}
\begin{cases}
f^{(\mathrm{in})}[\mathbf x + \mathbf y]
\cdot
g[\mathbf y]&
\text{if } \mathbf x+\mathbf y\in \mathcal D\\
0&
\text{otherwise,}
\end{cases}
\end{align}
where $\mathcal K$ is the kernel domain, such as $\{-1, 0, 1\}^2$ and $\mathcal D$ is the image domain in pixel coordinates from zero to image width and height, respectively. 
We extend this to \texttt{explicit} boundary handling $\ast_\mathrm e$ using a family of kernels $g_{1,\ldots n}$ as \begin{align}
f^{(\mathrm{out})}[\mathbf x]\ast_\mathrm e g_{1,\ldots,n}
=
\sum_{\mathbf y \in \mathcal K}
\begin{cases}
f^{(\mathrm{in})}[\mathbf x + \mathbf y]
\cdot
g_{s[\mathbf x]}[\mathbf y]& 
\text{if } \mathbf x+\mathbf y\in \mathcal D\\
0&
\text{otherwise,}
\end{cases}
\end{align}
where $s[\mathbf x]$ is a selection function that returns the index from 1 to $n$ of the filter to be used at position $\mathbf x$ (\refFig{Domain}).
The number of filters $n$ depends on the size of the receptive field: For a 3$\times$3 filter it is 9 cases, for larger fields it is more.

\paragraph{Loss}
The loss is defined on multiple filter kernel values $g_{1,\ldots n}$ instead of a single kernel. 
As this construction comprises of linear operations only (the selection function can be written as nine multiplications of nine convolution results with nine masks that are 0 or 1 and a final addition), it is back-propagatable.


\paragraph{Implementation}
A few things are worth noting for the implementation.
First, applying multiple kernels in this fashion has the same complexity as applying a single kernel.
Convolution in the Fourier domain, where costs would differ, is typically not done for kernels of this size.
Second, the memory requirement is the same as when running with common boundary conditions.
All kernels jointly output one single feature image.
The boundary filters are never run and no result is stored at the interior. 
The only overhead is in storing the filter masks.
In practice however, implementation constants might differ between implementations, in particular for parallel machines (GPUs).

The first practical option for implementation is the most \emph{compatible} one that just performs all nine convolutions on the entire image and later composes the nine images into a single image.
This indeed has compute and memory cost linear in the number of filters, \ie nine times more expensive, both for training and deployment

To avoid the overhead, without having to access the low level code of the framework in use, the additional kernels can be trained on the specific sub-parts of the input that they act on and then composited back to form the output.



\mysection{Analysis}{Analysis}
We will now analyze the effect of border handling for a simplified task and different networks: learning how to perform a Gaussian blur of a fixed size.
Despite the apparent simplicity, we will see, how many different variants of a state-of-the-art U-net-like \cite{ronneberger2015u} architecture all suffer from similar boundary handling problems.
This indicates, that the deteriorating effect of unsuccessful boundary handling cannot be overcome by adapting the network structure, but needs the fundamentally different domain decomposition we suggest. 

\mysubsection{Methods}{Methods}

\paragraph{Task}
The tasks is to learn the effect of a Gauss filter of size 13$\times$13 to 128$\times$128 images, obtained from the dataset used for the ILSVRC \cite{ILSVRC15} competition, comprising of over one million images selected from ImageNet \cite{Deng09imagenet}.
The ground truths were computed over $128+12 \times 128+12$ images, which were then cropped to 128$\times$128.

\paragraph{Metrics}
We compare to the reference by means of the MSE metric, which was also used as the loss function.
The models were selected by comparing the loss values over validation set, while the reported loss values were separately computed over a test set comprising of 10\,k examples.

\paragraph{Architecture}
We use a family of architectures to cover both breadth and width of the network.
The breadth is controlled by the number of feature channels and the depth by the number of layers.
More specifically, the architecture comprises of $n_\mathrm l$ layers.
Each layer performs a convolution to produce $n_\mathrm f$ feature channels, followed by a ReLU non-linearity.
We choose such an architecture, to show that the effect of boundary issues is not limited to a special setting but remains fundamental.

\paragraph{Boundary handling}
We include our \texttt{explicit} handling, as well as the classic \texttt{zero} strategy that assumes the image to be 0 outside the domain and \texttt{reflect} padding, that reflects the image coordinate around the edge or corner.

\mysubsection{Experiments}{AnalysisExperiments}
Here, we study how different architecture parameters affect boundary quality for each type of boundary handling.

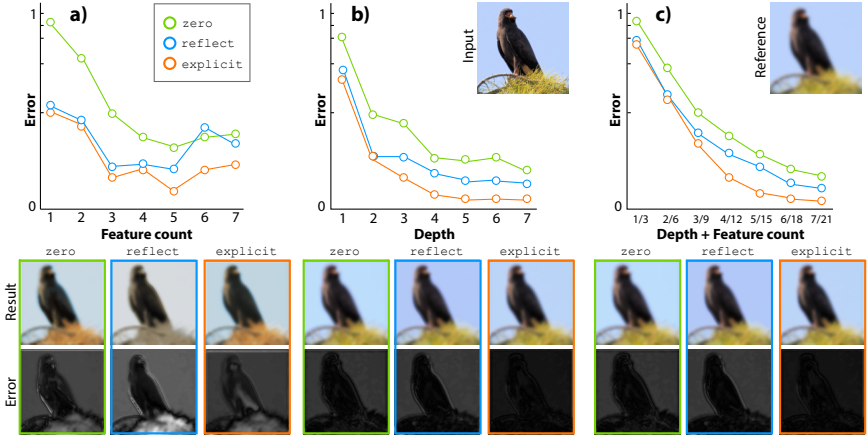
\myfigure{Analysis}{Analysis of different architectural choices  using different boundary handling (colors).
First, \textbf{(a)} we increase feature channel count (first plot and columns of insets).
The vertical axis shows log error for the MSE (our loss) and the horizontal axis different operational points.
Second \textbf{(b)}, depth of the network is increased (second plot and first 4 columns of insets).
Third, \textbf{(c)} both are increased jointly.
The second row of insets shows the best (most similar to a reference) result for each boundary method (column) across the variation of one architectural parameter for a random image patch (input an reference result seen in corner).}

\paragraph{Varying depth}
When varying depth $n_\mathrm l$ from a single up to 7 layers (\refFig{Analysis}a) we find, that our \texttt{explicit} boundary handling performs best on all levels, followed by \texttt{reflect} boundary handling and \texttt{zero}.
The feature channel count is held fixed at $n_\mathrm f=3$.

\paragraph{Varying feature count}
When varying feature channel count $n_\mathrm f$, it can be seen that \texttt{explicit} leads the board, followed by \texttt{reflect} and \texttt{zero} (\refFig{Analysis}b).
The depth is held fixed at $n_\mathrm d=2$.

\paragraph{Varying feature count and depth}
When varying both depth $n_\mathrm l$ and feature channel count $n_\mathrm f$, seen in \refFig{Analysis}, c we find, that again no architectural choice can compensate for the boundary effects.
Each of the seven steps increase feature count by 3 and depth by 1.

\paragraph{Statistical analysis}
A two-sided $t$ test ($N=10,000$) rejects the hypothesis that our method is the same as any other method for any task with $p<.001$. 

\mysection{Applications}{Applications}
Now, we compare different boundary handling methods in several typical applications.

\mysubsection{Methods}{Methods}

\paragraph{Architecture}
We use an encoder-decoder network with skip connections \cite{ronneberger2015u} optimized for the MSE loss using the ADAM optimizer
.
Details are shown in our supplemental materials.
The architecture is different from the simplified one in the previous section where it was important to systematically explore many possible variants.
The encoding proceeds in $3\times 3$ convolution steps 1 to 7, increasing the number of feature channels from 1 to 256.
There is a flat 1$\times$1 convolution at the most abstract representation at stage 8.
Decoding happens on stages 9 to 14.
This step resizes the image, convolves with stride 1 and outputs the stated number of feature, followed by a concatenate convolution by the stated skip ID and finally a convolution with stride 1 that outputs the stated number of features (\texttt{ResConv}).

Note, that boundary handling is required at all stages except 8.
For the down-branch 1--7 this can be less relevant as strides do not produce all edge cases we handle, \eg the boundary pixels on the bottom of the input are skipped in an even resolution scheme. 

\paragraph{Measure}
We apply different task-specific measures: Gauss filtering and Colorization produce images for human observers and consequently are quantified using DSSIM.
De-Bayering, as a de-noising task, is measured using the PSNR metric while disparity and scene flow are image correspondence problems with results in pixel units.

Additionally, we propose to measure the success as the \emph{loss ratio} between the test loss of our architecture with and the test loss of an architecture without explicit boundary handling, using the MSE metric.
We suggest to use the ratio as it abstracts away from the unit and the absolute loss value that depends on the task, allowing to compare effectiveness across tasks.

\mysubsection{Results}{ApplicationResults}

\addtolength{\tabcolsep}{0.00cm}  
\begin{table}
\small
\center
\caption{
Quantitative results.
Different rows are different tasks.
Different columns express different measures and different methods.
\emph{Absolute error} is measured using different metrics (eventually not identical to the loss), while the \emph{Error ratio} is expressed as the ratio of the loss of our method over the opposing boundary rules.
Best is bold.
}
\label{tbl:Main}
\begin{tabularx}{\textwidth}{rlr ccc ccc}
&&
\multicolumn{4}{c}{Absolute error}&
\multicolumn{3}{c}{Error ratio}
\\
&&
\multicolumn{4}{c}{Other metric}&
\multicolumn{3}{c}{MSE ratio}
\\
\tblhead{Task}&
\tblhead{Src}&
\tblhead{Unit}&
\tblhead{\texttt{refl}}&
\tblhead{\texttt{zero}}&
\tblhead{Ours}&
\tblhead{\texttt{refl}}&
\tblhead{\texttt{zero}}&
\tblhead{Ours}
\\
\toprule
Gauss filtering& & 
DSSIM &.0018& .0022& \textbf{.0016}& 79\,\%& 83\,\%& \textbf{100}\,\%\\
De-noise/Bayer& \cite{gharbi2016deep}&
PSNR& 31.46& 31.50& \textbf{31.94}& 90\,\%& 89\,\%& \textbf{100}\,\%\\
Colorization& \cite{zhang2016colorful}&
DSSIM & .1593& .1604& \textbf{.1577}& 99\,\%& 98\,\%& \textbf{100}\,\%\\
Disparity& \cite{dosovitskiy2015flownet}&
px & 1.538& 1.511& \textbf{1.403}& 84\,\%& 88\,\%& \textbf{100}\,\%\\
Scene flow& \cite{dosovitskiy2015flownet}&
px  & 1.380& 1.183& \textbf{1.096}& 56\,\%& 73\,\%& \textbf{100}\,\%\\
\bottomrule
\end{tabularx}
\end{table} 

\paragraph{Gauss blur}
Gauss filtering is a simple baseline task with little relevance to any practical application as we know the solution (\refSec{Analysis}.)
It is relevant to our exposition, as we know that, if the network had seen the entire world (and not just the image content) it would be able to solve the task.
It is remarkable, that despite the apparent simplicity of the task -- it is a single linear filter after all -- the absolute loss is significant enough to be visible for classic boundary handling.
It is even more surprising, that the inability to learn a simple Gauss filter does not only result in artifacts along the boundaries, but also in the interior.
This is to be attributed to the inability of a linear filter to handle the boundary.
In other words, a network without explicit boundary handling is unable to learn a task as easy as blurring an image.
We will see that this observation can also be made for more complex tasks in the following sections.

\paragraph{De-noising and De-bayering}
In this application we learn a mapping from noisy images with a Bayer pattern to clean images using the training data of Gharbi~\etal \cite{gharbi2016deep}.
The measure is the PSNR, peak signal-to-noise ratio (more is better).
We achieve the best PSNR at $31.94$, while the only change is the boundary handling.
In relative terms, traditional boundary handling can achieve only up to $90\,\%$ of MSE.

\paragraph{Colorization}  
Here we learn the mapping from grey images to color images using data from Zhang\etal\cite{zhang2016colorful}.
The metric again is DSSIM.
We again perform slightly better in both absolute and relative terms.

\paragraph{Disparity and scene flow}
Here we learn the mapping from RGB images to disparity and scene flow using the data from Dosovitskiy\etal\cite{dosovitskiy2015flownet}.
We measure error in pixel distances (less is better).
Again, adding our boundary handling improves both absolute and relative error.
In particular, the error of \texttt{reflect} and \texttt{zero} is much higher for scene flow.

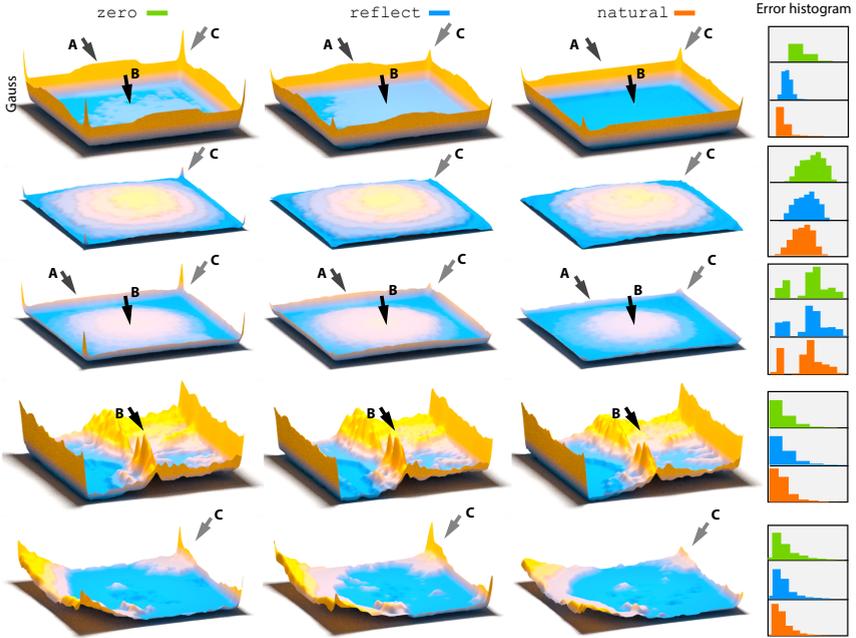
\myfigure{MeanErrors}{Mean errors across the corpus visualized as height fields for different tasks and different methods.
Each row corresponds to one task each column to one way of handling the boundary.
Arrow A marks the edge that differs (ours has no bump on the edge).
Arrow B mark the interior that differs (ours is flat and blue, others is non-zero, indicting we improve also inside).
Arrow C shows corners, that are consistently lower for us.
}

\mysection{Discussion}{Discussion}
We now will discuss the benefit and challenges of \texttt{explicit} boundary handling.

\paragraph{Overhead}
Here we study four implementation alternatives for \refSec{OurApproach}.
They were implemented as a combination of OpenGL geometry and fragment shaders.
The test was ran on a Nvidia Gefore 480, on a 3 mega-pixel image and a 3$\times$3 receptive field.

The \emph{first} method uses a simple \texttt{zero}-padding provided by OpenGL's \texttt{sampler2D}, invoking the GS once to cover the entire domain and applying the same convolution everywhere.
This requires 2.5~ms.
This is an upper bound for any convolution code.

The \emph{second} implementation executes nine different convolutions, requiring 22.5~ms.
This invokes the GS nine times, each invoking all pixels.

The \emph{third} variant invokes the GS once and a conditional statement for all pixels selects the kernel weights per-pixel in the domain.
This requires 11.2~ms.

The \emph{fourth} variant, a domain decomposition, invokes the GS nine times to draw nine quads that cover the respective interior and all boundary cases as seen in recall \refFig{Domain}.
Even after averaging a high number of samples, we could not find evidence for this to be slower than the baseline method \ie 2.5~ms.
This is not unexpected, as the running time for a few boundary pixels is below the variance of the millions of interior pixels.

In practice, the learning is limited by other factors such as disk-IO.
Our current implementation in Keras \cite{chollet2015keras}, offers a simple form of domain decomposition. We tested the performance loss over epochs with an average duration of 64 seconds. Our method results in a 0.2\.\% average performance loss over the classic \texttt{zero} rule. 

\paragraph{Scalability in receptive field size}
For small filters, the number of cases is small, but grows for larger filters.
Fortunately, the trend is to rather cascade many small filters in deeper network, instead of shallower networks with large filters.

\paragraph{Structure}
Here we seek to understand where spatially in the image the differences are strongest.
While our approach changes the processing on edges, does it also affect the interior?
We compute the per-pixel MAE and average this over all images in the corpus.
The resulting error images are seen in \refFig{MeanErrors}.
We found the new method to consistently improve results in the interior regions. It looks as if the new boundary rules effectively “shield” the inner regions from spurious boundary influences. The results at the boundaries are very competitive too, often better than \texttt{zero} and \texttt{reflect} boundary handling.
Note, that it is not expected for any method, also not ours, to have a zero error at the boundary: this would imply we were able to perfectly predict unobserved data outside of the image.

\begin{wrapfigure}[11]{l}{4cm}
\vspace{-0.1in}
\includegraphics[scale=0.6]{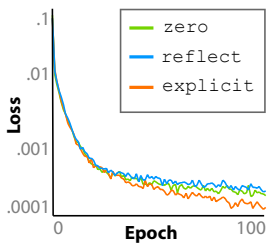}
\caption{Convergence rate with different types of  boundary handling.}
\label{fig:Loss}
\end{wrapfigure}

\paragraph{Convergence}
Convergence of both our approach and  traditional \texttt{zero} boundary handling is seen in \refFig{Loss}.
We find, that our method is not only resulting in a smaller loss, but also does so at the same number of epochs.
Before we have established that the duration of epoch are the same for both methods.
We conclude there is no relevant training overhead for our method.

\paragraph{Practical alternatives}
There are simpler alternatives to handle boundaries in an image of $n_\mathrm p$ pixels.
We will consider a 1D domain as an example here.
The first is to crop $n_\mathrm c$ pixels on each side and compute only $n_\mathrm p-2n_\mathrm c$ output pixels.
The cropping $n_\mathrm c$ is to be made sufficiently large, such that no result is affected by a boundary pixel and $n_\mathrm c$ depends on the network structure.
In a single-resolution network of depth $n_\mathrm d$ with a receptive field size of $2n_\mathrm r+1$, we see, that $n_\mathrm c=n_\mathrm d\times n_\mathrm r$.
In a multi-resolution network however, the growth is exponential, so $n_\mathrm c=n_\mathrm r^{n_\mathrm d}$, and for a typical encoder-decoder that proceeds to a resolution of $1\times 1$, every pixel is affected.
 This leaves two options: either the minimal resolution is capped and the CNN is applied in a sliding window fashion \cite{ronneberger2015u}, computing always only the unaffected result part, incurring a large waste of resources, or the network simply has to use its own resources to make do with the inconsistent input it receives.

\section{Conclusion}
In traditional image processing, the choice of boundary rule was never fully satisfying.
In this work, we provide evidence, that CNNs offer the inherent opportunity to jointly extract features and handle the boundary as if the image continues naturally.
We do this by learning filters that are executed on the boundary along with traditional filters executed inside the image.
Incurring little learning and no execution overhead, the concept is simple to integrate into an existing architecture, which we demonstrate by increased result fidelity for a typical encoder-decoder architecture on practical CNN tasks.

\section{Acknowledgements}
We thank Paul Guerrero, Aron Monszpart and Tuanfeng Yang Wang for their technical help in setting up and fixing the machines used to carry out the experiments in this work. This work was partially funded by the European Union's Horizon 2020 research and innovation programme under the Marie Sk\l{}odowska-Curie grant agreement No 642841, by the ERC Starting Grant SmartGeometry (StG-2013-335373), and by the UK Engineering and Physical Sciences Research Council (grant EP/K023578/1).


\bibliographystyle{bmvc2k_natbib}
\bibliography{paper}
\end{document}